\documentclass[letterpaper,conference]{IEEEtran}

\usepackage[english]{babel}
\usepackage{amsmath}
\usepackage{graphicx}
\usepackage{algorithm}
\usepackage{algorithmic}
\usepackage{paralist}
\usepackage{hyperref}

\newcommand{\figwidth}{0.47}

\newcommand{\beq}{\begin{equation}}
\newcommand{\eeq}{\end{equation}}
\newcommand{\bea}{\begin{eqnarray}}
\newcommand{\eea}{\end{eqnarray}}

\newcommand\fig[1]{Figure~\ref{fig:#1}}
\newcommand\tab[1]{Table~\ref{tab:#1}}
\newcommand\eqn[1]{Eq.~(\ref{eq:#1})}

\newcommand\alg[1]{Algorithm~\ref{alg:#1}}

\begin{document} 

\title{Dictionary Learning with\\Equiprobable Matching Pursuit}

\author{\IEEEauthorblockN{Fredrik Sandin}
\IEEEauthorblockA{EISLAB, Lule{\aa} University of Technology, Sweden\\
Email: fredrik.sandin@ltu.se}
\and
\IEEEauthorblockN{Sergio Martin-del-Campo}
\IEEEauthorblockA{SKF--LTU University Technology Center and\\
EISLAB, Lule{\aa} University of Technology, Sweden}
}

%\author{Fredrik Sandin and Sergio Martin-del-Campo%
%\thanks{F. Sandin is with  EISLAB, 
%Lule{\aa} University of Technology (LTU),
%97187, Lule{\aa}, Sweden (e-mail: fredrik.sandin@ltu.se).}%
%\thanks{S. Martin-del-Campo is with the SKF--LTU University Technology Center and EISLAB, 
%Lule{\aa} University of Technology (LTU),
%97187, Lule{\aa}, Sweden.}}

\markboth{IEEE}{Dictionary Learning with Equiprobable Matching Pursuit}

\maketitle

%%%%%%%%%%

\begin{abstract}
Sparse signal representations based on linear combinations of learned atoms have been used to obtain state-of-the-art results in several practical signal processing applications.
Approximation methods are needed to process high-dimensional signals in this way because the problem to calculate optimal atoms for sparse coding is NP-hard.
Here we study greedy algorithms for unsupervised learning of dictionaries of shift-invariant atoms and propose a new method where each atom is selected with the same probability on average, which corresponds to the homeostatic regulation of a recurrent convolutional neural network.
Equiprobable selection can be used with several greedy algorithms for dictionary learning to ensure that all atoms adapt during training and that no particular atom is more likely to take part in the linear combination on average.
We demonstrate via simulation experiments that dictionary learning with equiprobable selection results in higher entropy of the sparse representation and lower reconstruction and denoising errors, both in the case of ordinary matching pursuit and orthogonal matching pursuit with shift-invariant dictionaries.
Furthermore, we show that the computational costs of the matching pursuits are lower with equiprobable selection, leading to faster and more accurate dictionary learning algorithms.
\end{abstract}

%%%%%%%%%%%%%%%%%%%%%%%%%%%%%%%%%%%%%%%%%%%%%%%%%%%%%%%%%%%%%%%%%%%%%%%%%%%%%%%%%%%%%%%%%%%%%%%%%%%%%%%%%%%%%%%%%%%%%%%%%%%%%%%%%%%%%

%\begin{IEEEkeywords}
%dictionary learning, sparse approximation, matching pursuit, unsupervised learning, homeostatic regulation, neuromorphic engineering
%\end{IEEEkeywords}

%%%%%%%%%%%%

\section{Introduction}
Sensors like cameras, microphones and accelerometers typically generate redundant information because the resulting data have an underlying structure.
That is why most observed phenomena can be accurately approximated with mathematical relationships in the form of physical laws, and also why such data can be compressed using algorithms that identifies and approximates patterns.
Redundancies in the data make the processes of communicating, analyzing and storing information inefficient, thus constraining the domain of feasible applications. 
The problem to identify the structure of signals and thereby derive succinct representations that are both compact and informative is a computational challenge, which limits the efficacy of sensor systems.

Sparse representation models \cite{Mallat2008,Bruckstein2009,elad2010sparse,Nam2013} % Starck2010,
and related machine learning algorithms \cite{Rubinstein2010,Tosic2011} have proven remarkably successful at extracting useful information from complex high-dimensional signals, for example in the context of
denoising \cite{Blumensath2008}, % e.g. in paper by blumensath & davies , \cite{Blumensath2008,Blumensath2009}.
under-determined source separation \cite{Zibulevsky01,Bofill01},
compressed sensing \cite{Donoho06CS},
super-resolution sensing \cite{CPA21455},
and classification \cite{NIPS2008_3448}.
In a sparse model, the signal is typically described as a linear combination of elementary functions called atoms, which can either be predefined (if an appropriate generative model for the signal class is known) or learned from a training signal.
The goal is to select or learn the atoms so that the model residual is minimum for a certain sparsity of the representation, or to maximize the sparsity for a certain tolerance on the model residual.
Atoms typically have unit norm and the set of atoms defines a dictionary, which can be subject to further constraints like temporal or spatial translation invariance.

Fourier and wavelet analysis are two examples where predefined dictionaries are used.
Models based on predefined dictionaries have enabled derivation of closed-form mathematical results and fast algorithms that are widely used, but such approaches are simplistic compared to the complexity of the underlying natural phenomena.
The dictionary learning approach is based on the hypothesis that complex signals can be more accurately modeled by extracting environmentally matched atoms from the signal.
However, the dictionary learning problem is NP-hard and it is also hard to find approximate solutions near the optimal sparsity level \cite{Tillmann2014}.

The development of dictionary learning methods was stimulated by results presented in the mid '90s by Olshausen and Field \cite{Olshausen96,Olshausen97}, which demonstrate that atoms similar to the receptive fields of cells in visual cortex can be learned from natural images by imposing a few general optimization conditions, including sparsity and statistical independence of atoms.
This demonstrates that some aspects of the low-level functions in the visual system can be explained by a few general computational principles, and that elementary structures of such complex signals can be automatically uncovered from examples.
Since that time several probabilistic dictionary learning and sparse coding methods have been developed
\cite{Bruckstein2009,Rubinstein2010}, % ADD RECENT REFS HERE, including papers by Skretting & Engan
aiming for a dictionary that either maximizes the likelihood of the data,
as for example in \cite{Lewicki2000},
or the posterior probability of the dictionary,
as in \cite{KreutzDelgado2003}.
Recent developments include extensions of dictionary learning methods to distributed systems \cite{Chouvardas2015} and low-power hardware \cite{TNANO2015}.

There is a knowledge gap between receptive field models and the observed function of neural networks in biological sensory systems.
For instance, neurons demonstrate a form of homeostatic adaptation of the functional properties of the network to the ongoing changes in the statistical structure of the sensory input \cite{Fournier2011,Ramirez2014,Fournier2014}, which may be related to optimal encoding \cite{Wainwright1999} of the input by dynamic adaptation of the receptive fields.
Furthermore, there is a notion that homeostatic mechanisms serve to maintain the dynamics of cortical networks at a critical point \cite{Beggs2008,Beggs2012} where the dynamic range and information processing capacity are optimal \cite{Shew01022013,PhysRevE.90.062714}.
Can we learn something new about dictionary learning by incorporating and studying such homeostatic regulation mechanisms?

In this paper we focus on greedy algorithms for dictionary learning, where the approximation error is reduced iteratively by sparse decomposition of the signal followed by gradient optimization of the atoms in the dictionary.
Our aim is to study a basic implementation of homeostatic regulation where all atoms are enforced to occur with the same probability on average, which in terms of neurons and receptive fields mean that neurons fire with the same average probability.

\begin{figure}[b!]
\centering
\advance\leftskip-0.52cm
\includegraphics[width=0.54\textwidth]{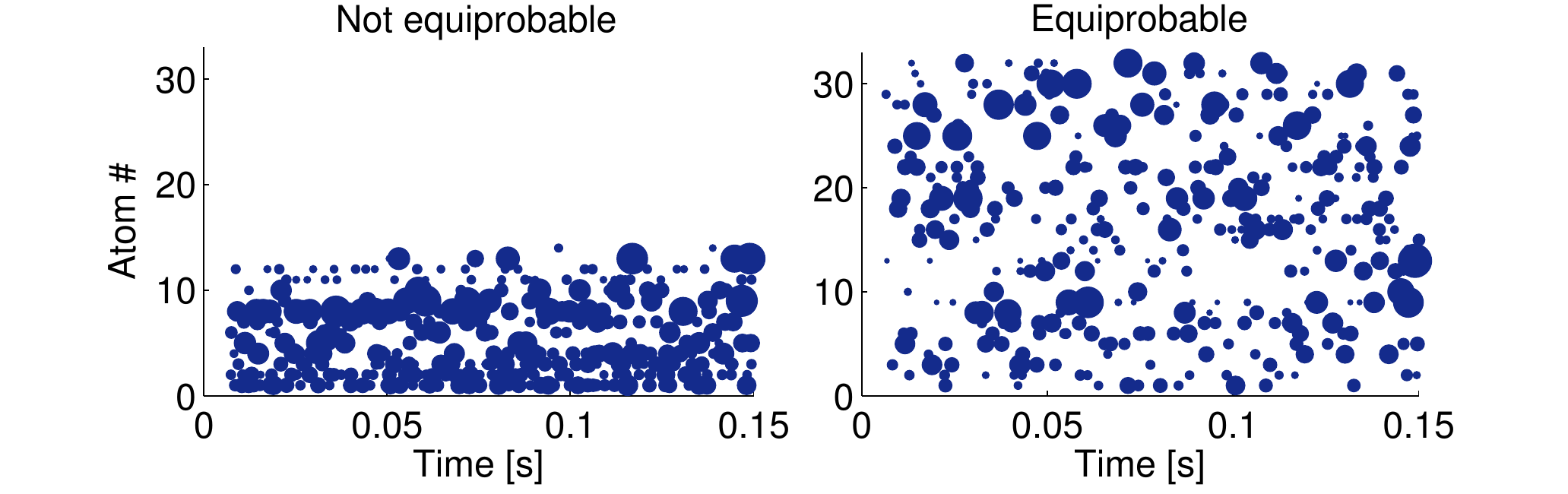}
\vspace{-0.6cm}
\caption{Two equally sparse representations of $0.15$~seconds of $44.1$~kHz rock music
obtained with two different matching pursuit algorithms and learned dictionaries.
Discs represent onsets of atoms and disc sizes are proportional to the atom coefficients.
The panel on the (left-hand) right-hand side is obtained with {(non-equiprobable)} equiprobable MP.
With equiprobable MP all atoms adapt to the signal and contributes to the sparse representation.
}
\label{fig:scatter}
\end{figure}

The starting point is the dictionary learning method introduced by Smith and Lewicki \cite{smith2006}, where the sparse code is generated with Matching Pursuit (MP) \cite{Smith2005a,Mallat1993} and the shift-invariant dictionary is updated with probabilistic gradient ascent on the likelihood of the data \cite{Lewicki2000}.
In principle this model corresponds to a particular type of recurrent convolutional neural network with max pooling (cf. Fig. 5 in \cite{Olshausen97}), which we extend here with a homeostatic regularization mechanism.
Typically a subset of the atoms are selected in the MP for one particular signal, while some atoms are rarely selected and do not take significant part in the gradient-based dictionary update process.
We implement homeostatic regulation by enforcing equiprobable atom selection in the MP, which implies that each atom is equally likely to occur in the linear combination on average, see \fig{scatter}.
In addition to MP we consider dictionary learning with Orthogonal MP (OMP) \cite{Mallat1993,Pati1993}, which in the case of shift-invariant dictionaries is applicable to high-dimensional signals in the form of local OMP \cite{LocOMP2009}.

\begin{figure}[ht!]
\centering
\vspace{-0.5cm}
\advance\leftskip-0.14cm
\includegraphics[width=0.46\textwidth]{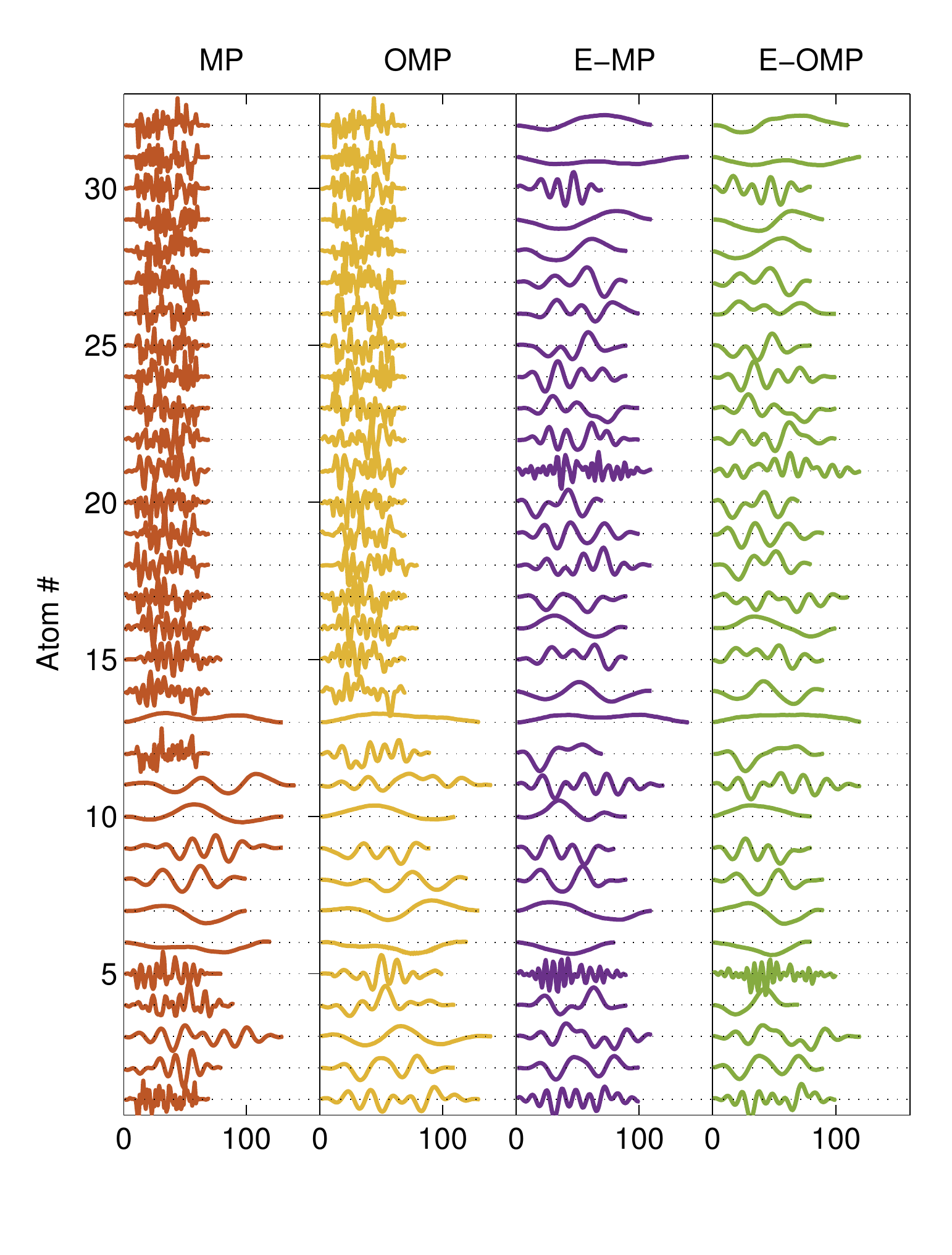}
\vspace{-0.8cm}
\caption{Dictionaries of 32 atoms learned from $44.1$~kHz rock music using the dictionary learning methods described below, which are based on four different matching pursuit algorithms (MP, OMP, E--MP and E--OMP).
Initially the four dictionaries are identical and include atoms that are sampled from a Gaussian distribution with zero mean.
}
% Dictionaries obtained by training on 3000s of 44.1kHz audio in song1.mp3.
\label{fig:dict32}
\end{figure}

We study the effects of equiprobable selection on the learned dictionary and sparse representation of music and birdsong.
With equiprobable selection all atoms adapt to the training signal, see \fig{dict32}.
Our main finding is that the entropy and reconstruction accuracy are higher for equiprobable sparse representations, and that the computational cost is reduced by the equiprobability constraint.
Equiprobable selection can be generalized to other greedy dictionary learning methods and these results motivate further investigations of dictionary learning models incorporating more realistic homeostatic regulation mechanisms observed by neuroscientists.

%%%%%%%%%%%

%%%%%%%%%%%

\section{Dictionary learning method}
\label{sec:sparse}

The signal, $x(t)$, is modelled as a linear superposition of waveforms, $\varphi_i$, with compact support and additive noise
\beq
    x(t) =  \sum_{i=1}^M \sum_{j=1}^{N_i} a_{i,j} \varphi_{i}(t - \tau_{i,j}) + \epsilon(t).
    \label{eq:genmodel}
\eeq
The functions $\varphi_i$ are shift-invariant {\em atoms} that represent elementary waveforms of the signal and $M$ is the number of different atoms in the model.
The variable $\epsilon(t)$ represents the model residual, including noise.
The variable $N_i$ refers to the number of instances of atom $\varphi_i$,
and the temporal position and amplitude of the $j$-th instance of atom $\varphi_i$ are denoted by $\tau_{i,j}$ and $a_{i,j}$, respectively.
The set of  $M$ atoms defines a dictionary
\beq
\Phi = \left\{ \varphi_1, \cdots, \varphi_{M} \right\},
\eeq
which we want to adapt to the signal.
In principle $x(t)$ and $\varphi_i$ can be multidimensional, but here we limit the numerical experiments to scalar signals.

\eqn{genmodel} defines a sparse approximation of $x(t)$ if the number of terms are few compared to the number of samples of $x(t)$ and $\epsilon(t)$ is small compared to $x(t)$,
which requires that the dictionary $\Phi$ is adapted to the signal.
Thus, the dictionary learning problem is to calculate a set of $\varphi_i$ that minimizes $\epsilon(t)$ under $\ell_0$ regularization of the coefficient matrix.
This problem is NP-hard \cite{Tillmann2014} and cannot be solved explicitly.
Instead, a greedy algorithm that reduces the approximation error in a two-step iteration process is used \cite{Lewicki1999}:
\begin{inparaenum}[\itshape a\upshape)]
	\item[$A$)] \textit{Encoding step}; optimize the sparse representation of the signal $x(t)$ with MP or OMP and a constant dictionary $\Phi$.
    \item[$B$)] \textit{Learning step}; update the dictionary $\Phi$ using the sparse representation and model residual so that the approximation error is reduced.
\end{inparaenum}

The encoding step is by itself an iterative process that terminates at some predefined sparsity or tolerance on the model residual.
Each iteration of the encoding process includes two steps:
\begin{enumerate}
	\item[$A_1$)] \textit{Atom selection} -- Find the atom $\varphi_i$ and offset $\tau_{i,j}$ that maximizes the cross-correlation with the model residual, $ r_{k}(t)$, and calculate the corresponding $a_{i,j}$.
    \item[$A_2$)] \textit{Residual update} -- Update the residual by subtracting the contribution from the selected atom, $r_{k+1}(t) = r_{k}(t)-a_{i,j}\varphi_{i}(t-\tau_{i,j})$ with $r_{0}(t)=x(t)$.
\end{enumerate}
The accuracy of the resulting signal approximation depends on the complex interplay between the atom selection rule and the dictionary learning process.

MP optimizes the parameters $\tau_{i,j}$ and $a_{i,j}$ of the most recently selected atom, while local OMP \cite{LocOMP2009} re-optimizes all $a_{i,j}$ for selected atoms with overlapping support in each iteration (OMP compensates for the interference between atom instances).
In the following we consider both MP and OMP when introducing homeostatic regulation.
Our implementation of MP and OMP are based on efficient computational methods like those described in \cite{LocOMP2009} and \cite{Krstulovic2006}.

%\subsection{Equiprobable matching pursuit}

The atom selection rule defined above is optimal for sparse decomposition of a signal with a constant dictionary \cite{Mallat1993}, but it does {\it not} imply optimal dictionary  learning.
For example, if some atoms are more frequently selected than others the entropy of the resulting sparse representation is expected to be suboptimal, potentially leading to information loss and lower reconstruction accuracy.
Furthermore, atoms that are rarely selected  mainly contribute to the model complexity. 
Therefore, we introduce a modified atom selection rule that regulates the average probability for each atom to be selected:
\begin{enumerate}
	\item[$A^*_1$)] \textit{Equiprobable atom selection} -- Find the atom $\varphi_i$ and offset $\tau_{i,j}$ that maximizes the cross-correlation with the model residual, $ r_{k}(t)$, under the constraint $P(\varphi_i) < p$ and calculate the corresponding $a_{i,j}$.
\end{enumerate}
The atom selection constraint $P(\varphi_i) < p$ can be applied to both MP and OMP for dictionary learning purposes.
The constrained matching pursuit terminates when no atom is selected, at which point all atoms occur with the same probability $p$.
We refer to the resulting ``equiprobable'' MP and OMP as E--MP and E--OMP, respectively.
MP and OMP are defined by steps $A_1$, $A_2$ and $B$ above, while E--MP and E--OMP are defined by $A^*_1$, $A_2$ and $B$.
The resulting four matching pursuits are summarized in Table~I and Algorithm~1,
\begin{table}[b!]
\caption{List of matching pursuits defined by \alg{elocomp}.}
\label{tab:mp}
\begin{center}
\begin{small}
\vskip -0.1in
\begin{tabular}{|l|cc|}
\hline
\textsc{Method} 
& \texttt{constraint} 
& \texttt{neighborhood}
\\
\hline
\textsc{MP}    
	& ${\varphi_i\in\Phi}$ 
    & $\varphi_k$ \\
\textsc{OMP}
	& ${\varphi_i\in\Phi}$ 
    & $\Phi_k$ \\
\textsc{E--MP}
	&${\varphi_i\in\Phi, P(\varphi_i,\Phi_k) < p}$
    & $\varphi_k$ \\
\textsc{E--OMP}
	& ${\varphi_i\in\Phi, P(\varphi_i,\Phi_k) < p}$
	& $\Phi_k$ \\
\hline
\end{tabular}
\end{small}
\end{center}
\vskip -0.05in
\end{table}
\begin{algorithm}
\caption{Matching pursuit}\label{alg:elocomp}
\begin{algorithmic}[1]
	\FUNCTION{match$(\Phi, x, p)$}
	\STATE $r_0 = x$
	\STATE $a_0 = 0$
    \STATE $\tau_0 = 0$
	\STATE $\Phi_0 = \emptyset$
	\FOR{$k=1$ to $I$}
    	\STATE \{$\varphi_k,t_k\} = $~argmax$_{\varphi_i,\tau_{i,j}}\langle r_{k-1},\Phi\rangle$\\ ~~~~~~~~~~~~~~subject to \texttt{constraint}
        \STATE $\Phi_k = \Phi_{k-1} \cup \varphi_k$
        \STATE $\Psi_k =$~\texttt{neighborhood}
        \STATE $\chi_k = (\Psi_k^*\Psi_k)^{-1}\Psi_k^*r_{k-1}$
        \STATE $a_k = a_{k-1} + \chi_k$
        \STATE $\tau_k =\tau_{k-1} + t_k$
        \STATE $r_k = r_{k-1} - \Psi_k\chi_k$
	\ENDFOR
    \STATE \bfseries{return}~$\{a_I, \tau_I, r_I\}$
    \ENDFUNCTION
\end{algorithmic}
\end{algorithm}
which is a straightforward extension of the local OMP algorithm presented in \cite{LocOMP2009} to equiprobable atom selection.

The function $P(\varphi_i,\Phi_k)$ is the probability for each shift-invariant atom to be selected, which is defined here by the relative number of occurrences of each atom in the subdictionary $\Phi_k$ of selected atoms and offsets.
For E--MP and E--OMP the total number of iterations, $I$, is defined so that $P(\varphi_i,\Phi_I) \equiv p$ for each shift-invariant atom, $\varphi_i$, which implies that each atom is included in the sparse approximation with the same probability, $p$, on average.
For MP and OMP the total number of iterations, $I$, is defined to be equivalent to the number of E--MP and E--OMP iterations.
Thus, the resulting MP, OMP, E--MP and E--OMP approximations are equally sparse and can be compared in terms of reconstruction error, denoising error etc.
With MP and OMP the shift--invariant atoms typically occur with different probabilities, resulting in a learned dictionary where only a subset of the atoms adapt to the training signal.

%%%%%%%%%%%%%%

%\subsection{Learning dictionary of shift-invariant atoms}
%\label{sec:learning}

The dictionary learning problem is to infer the set of atomic waveforms, $\varphi_i$, in the dictionary, $\Phi$, so that the matching pursuit results in a sparse representation with low residual.
A computationally feasible formulation of this problem can be obtained by rewriting \eqn{genmodel} in probabilistic form
\bea
        p(x | \Phi) &=& \int p(x | a,\Phi) p(a) da \label{eq:probrepr} \\
         &\approx& p(x | \hat{a},\Phi) p(\hat{a}), \label{eq:probreprS}
\eea
where $\hat{a}$ is the maximum a posteriori (MAP) estimation of $a$,
\beq
        \hat{a} = \arg \max_{a} p(a | x, \Phi) = \arg \max_{a} p(x | a, \Phi )  p(a),
        \label{eq:inferS}
\eeq
that is generated by the matching pursuit \cite{Lewicki2000,smith2006,Smith2005a}.
Furthermore, we assume that the noise term, $\epsilon(t)$, in \eqn{genmodel} is Gaussian.
Thus, the data likelihood, $p(x | a, \Phi )$, is also Gaussian and takes the form
\beq
        p(x | a, \Phi )  \approx \exp\left(- \frac{\| x-a\Phi \|^2}{2\sigma_{\epsilon}^{2}}\right)                   ,
        \label{eq:likeli}
\eeq
where
\beq
        \| x-a\Phi \|^2 = \| x - \sum_{i=1}^M \sum_{j=1}^{N_i} a_{i,j} \varphi_{i}(t - \tau_{i,j}) \|^2,
        \label{eq:likeliExp}
\eeq
and $\sigma_{\epsilon}^{2}$ is the variance of the noise.
Note that $x$, $a$ and $\Phi$ are matrices in these probabilistic expressions, and that the dictionary, $\Phi$, includes all possible shifts of each atom $\varphi_i$.
%
%The second assumption concerns the prior, $P(a)$, which is defined so that it promotes sparse representations and statistically independent components.
%Statistical independence is incorporated by
%\beq
%        P(a) =  \prod P(a_i),
%        \label{eq:prior}
%\eeq
%which means that the probability of a state $a$ is given by the product of individual probabilities $a_i$. % ????
%The probabilities $P(a_i)$ are modeled with a distribution that promotes sparseness (the Laplacian)
%\beq
%        P(a_i) \approx  \exp(-\theta |a_i| ).
%        \label{eq:priorI}
%\eeq
%However, it is important to note that sparseness is obtained via the encoding algorithm. % and that the prior does not affect the learning of the atoms. TRUE?????????
%Refer to Lewicki and Sejnowsli \cite{Lewicki2000} for details about the prior and the resulting sparse representation.

Under these assumptions the atoms can be optimized by performing gradient ascent on the approximate log data probability \cite{smith2006}.
It follows from \eqn{probreprS} that
\beq 
\frac{\partial}{\partial \varphi_i} \log (p(x | \Phi)) = \frac{\partial}{\partial \varphi_i} \left[ \log (p(x | \hat{a},\Phi)) + \log (p(\hat{a}))  \right].  \label{eq:probreprSder}
\eeq
By taking the derivative and substituting the likelihood term with \eqn{likeli} this becomes
\bea
        &&   \frac{\partial}{\partial \varphi_i} \log (p(x | \Phi)) \nonumber \\
        &=& \frac{-1}{2\sigma_{\epsilon}^{2}} \frac{\partial}{\partial \varphi_i} \| x - \sum_{i=1}^M \sum_{j=1}^{N_i} a_{i,j} \varphi_{i}(t - \tau_{i,j}) \|^2       \label{eq:probreprSderSolv}  \\
        &=&
	 \frac{1}{\sigma_{\epsilon}^{2}} \sum_j  a_{i,j}\left[r_I\right]_{\tau_{i,j}}.    \label{eq:probreprSderSolvsimp}
\eea
The factor $\left[r_I\right]_{\tau_{i,j}}$ represents the model residual coinciding with atom $\varphi_i$ at temporal position $\tau_{i,j}$.
In other words, the shape of each atom is adapted with a weighted average of the residual elements coinciding with the matches identified by the matching pursuit.
This is a form of nonlinear Hebbian learning because the atoms are adapted to patterns in the signal that they correlate with.

In order to use the gradient for dictionary learning we introduce a relative steplength parameter, $\eta$, and define the gradient ascent update of atom $\varphi_i$ as
\beq
        \varphi_i
        \to  \varphi_i + \frac{\eta}{\sigma_{\epsilon}^{2}} \sum_j  a_{i,j}\left[r_I\right]_{\tau_{i,j}}.    
        \label{eq:dlearn}
\eeq
This implies that the dictionary adaptation rate depends on the activation rate of atoms.
We zero-pad all atoms with ten elements and grow each tail in length with ten additional elements if the RMS of the tail exceeds 0.1 of the atom RMS.
The resulting dictionary learning method is summarized in Algorithm 2.

\begin{algorithm}
\caption{Dictionary learning}\label{alg:dl}
\begin{algorithmic}[1]
	\FUNCTION{dlearn$(M,p,\eta)$}
    \STATE $n = 0$
    \STATE $r_0 = 0$
    \STATE $\Phi^{(0)} = \texttt{randdict}(M)$
	\WHILE{$x_{n+1} = \texttt{getdata}(n, r_n)$ is not empty}
        \STATE $n = n + 1$
    	\STATE $\{a_n, \tau_n, r_n\} = \texttt{match}(\Phi^{(n-1)}, x_n, p)$
        \FOR{each $\xi_i$ in $a_n$}
        	\STATE $\delta_i = \eta~\xi_i [r_n]_{\tau_n}
            	/ \texttt{var}(r_n)$
        	\STATE $\varphi_i^{(n)} = \texttt{extnorm}
            (\varphi_{i}^{(n-1)} + \delta_i)$,
            ~~~$\varphi_{i}^{(n)} \in \Phi^{(n)}$
        \ENDFOR
    \ENDWHILE
    \STATE\bfseries{return}~$\Phi^{(n)}$
    \ENDFUNCTION
\end{algorithmic}
\end{algorithm}

The function $\texttt{randdict}(M)$ generates a random dictionary of $M$ normalized atoms, where each atom includes fifty elements sampled from a Gaussian distribution with zero mean and two vanishing tails that are ten elements long.
Thus, the $M$ different atoms are seventy elements long initially.
The $\texttt{extnorm}(\varphi)$ function extends the length of an atom {$\varphi$} with ten vanishing elements whenever the RMS of a tail exceeds the predefined threshold mentioned above, and it also normalizes the atom.
Training data is fetched with the $\texttt{getdata}()$ function, which allows block-based processing of sampled signals using overlapping window functions and part of the former residual, $r_{i-1}$, of the matching pursuit.

%%%%%%%%%%%%%%%%%%%%%%%%%%%%%%%%%%%%%%%%%%%%%%%%%%%%%%%%%%%%%%%%%%%%%%%%%%%%

%\begin{itemize}
%\item Apply to different signals: Vibration, Music, ...? Present summary in table.
%\item Event weight vs iteration number should be a staircase in the beginning, what about the end?
%\item Effect of dictionary size, 16 vs 32 etc?
%\item Cross validation shows better generalization ability?
%\item Scatter plots look different for ordinary and equiprobable pursuits? 
%\item Equiprobable pursuits allows for better scaling / hardware use?
%\item Classification accuracy?
%\item Forcing an equal number of events with short windows should be sub-optimal,
%an alternative is to introduce variable biases/scale factors in the selection step?
%\end{itemize}

\section{Results}

We investigate the effects of equiprobable atom selection on the learned dictionary and sparse approximation accuracy with numerical experiments using two different signals.
One signal is a 155~seconds long 44.1~kHz rock music track with lyrics \cite{FMA_rocksong}
and the second signal is a 26~seconds long 48~kHz recording of Zebra Finch song phrases \cite{Zebra_finch}.
The dictionary learning method is defined by \eqn{dlearn} with steplength $\eta=10^{-6}$.
We do not observe significant improvements in the resulting model accuracy using other values of $\eta$ and our qualitative analysis does not depend on finetuning of this hyperparameter.
The $\texttt{getdata}()$ function is defined so that each block of data ($x_n$ in Algorithm 2) is five seconds long sampled from one random location in the data set.
The progression of the dictionary learning protocol is thereby quantified in terms of time rather than epochs.

%%%%%%%%%%%%%%%%%%%%%%%%%%%%%%%%%%%%%%%%%%%%%%%%%%%%%%%%%%%%%%%%%%%%%%%%%%%%

\subsection{Model accuracy and rate of convergence}

\begin{figure}[t!]
\centering
\includegraphics[width=\figwidth\textwidth]{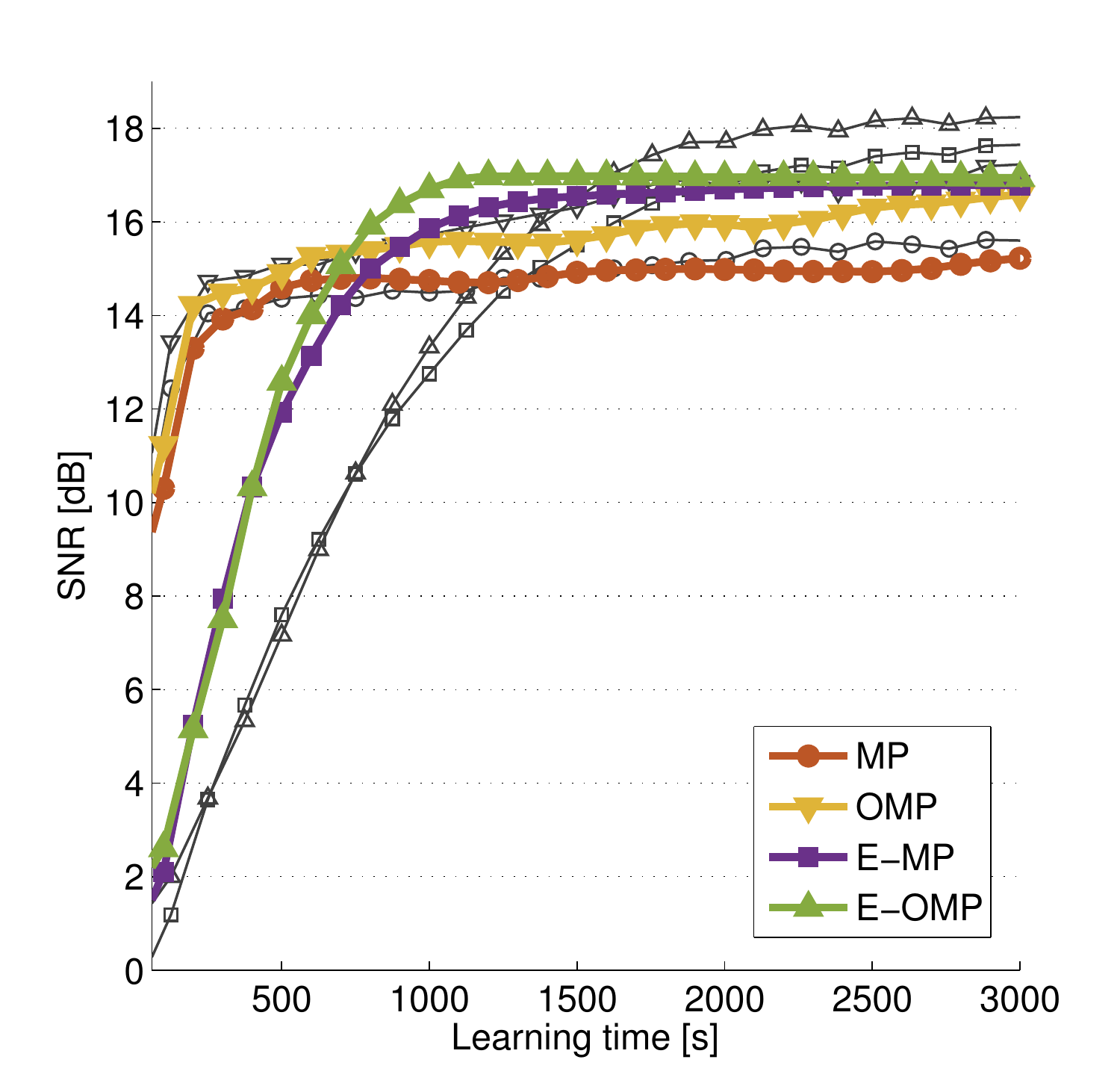}
\caption{Reconstruction SNR versus learning time for rock music with an average atom selection probability of $p=0.05$.
Bold (thin) lines are obtained with a dictionary of 32 (64) learned atoms.
}
% Training on a continuous and repeating 44.1kHz audio signal (song1.mp3).
\label{fig:recsnr}
\end{figure}

In the first experiment we study the signal-to-noise ratio (SNR) of the sparse approximation of rock music with an average atom selection probability of $p=0.05$, see \fig{recsnr}.
The resulting dictionaries of 32 atoms are displayed in \fig{dict32}.
Equiprobable selection reduces the initial dictionary learning rate,
but it also leads to an improved convergence time and accuracy of the sparse model.
For example, the accuracy of the E--MP-based approximation exceeds that of the OMP-based approximation after about 1000~seconds of learning.
This is remarkable considering that the computational cost of E--MP
is lower than both MP and OMP (further details below).
With 64 atoms a longer learning time is needed to reach a comparable SNR,
but after about 2000~seconds of learning the SNR exceeds that of models with 32 atoms.
Note that atoms are shift invariant.
Thus, the number of possible sparse representations of a five-second long window is astronomical even if there are only a few atoms in the dictionary.

\subsection{Effect of varying atom selection probability}
\begin{figure}[t!]
\centering
\includegraphics[width=\figwidth\textwidth]{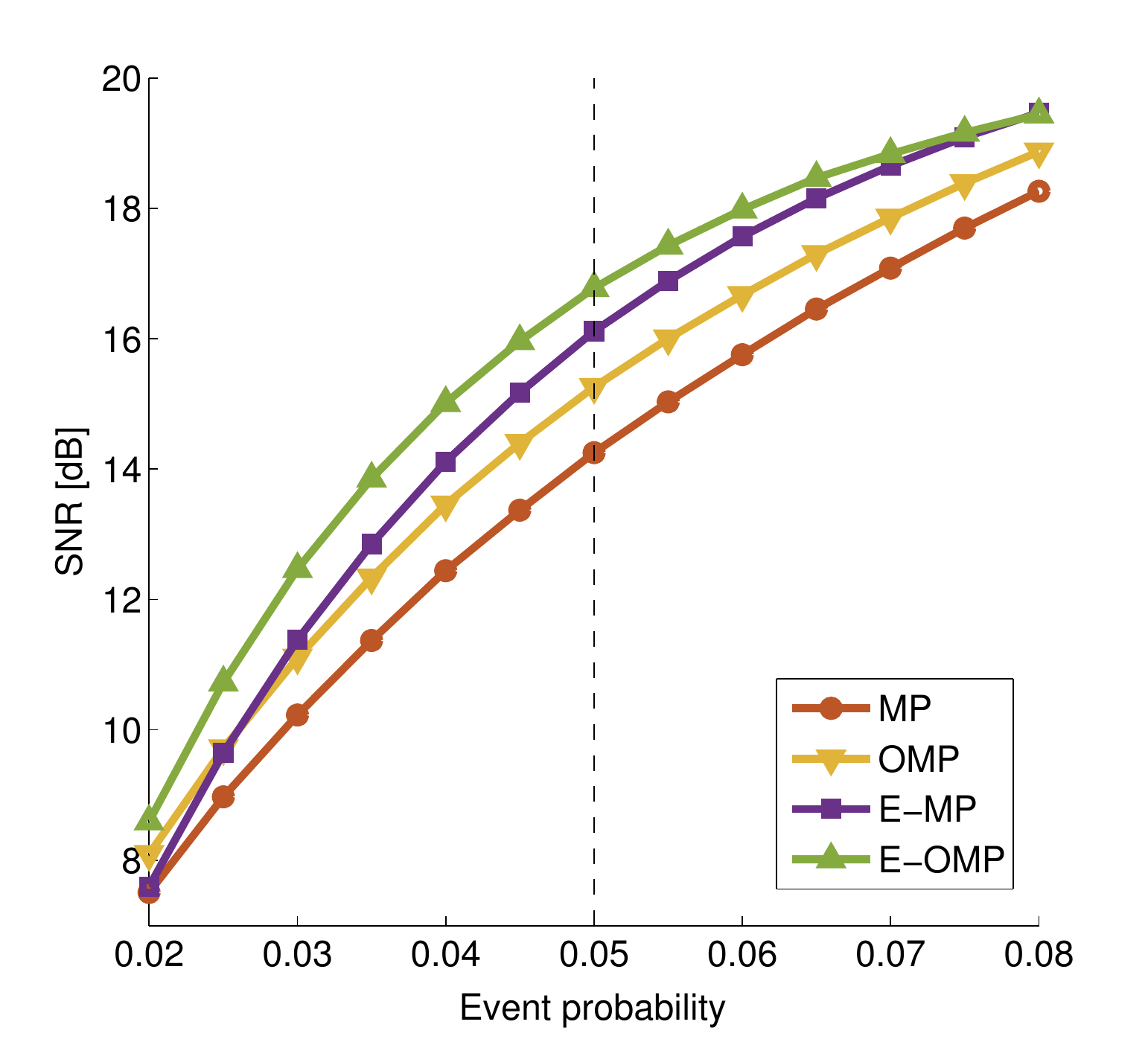}
\caption{{
Reconstruction SNR versus the average atom selection probability.
Dictionaries are learned from 1500~seconds of rock music with an average event probability of  0.05 (dashed line).}}
% Result for 5s of 44.1kHz audio data (song1.mp3 70-75s).
% A five second segment with both music and voice is selected.
% Dictionaries trained on 1500s of 44.1kHz audio (song1.mp3).
\label{fig:recacc}
\end{figure}

Next we study the sensitivity of the model to variations in the average atom selection probability, $p$.
Using dictionaries learned from 1500~seconds of rock music we calculate the SNR of the sparse approximation of rock music for different values of $p$, see \fig{recacc}.
This result demonstrates that the models based on dictionaries learned with E--MP and E--OMP degrades gracefully when the sparsity of the model changes.
The learned dictionaries generalize to such varying conditions and are not ``overfitted'' to one particular value of $p$.
Thus, equiprobable atom selection/activation can be feasible also with low-power computing substrates like neuromorphic chips where the average value of $p$ cannot be precisely defined due to device mismatch and noise.

%%%%%%%%%%%%%%%%%%%%%%%%%%%%%%%%%%%%%%%%%%%%%%%%%%%%%%%%%%%%%%%%%%%%%%%%%%%%

\subsection{Denoising}

\begin{figure}[t!]
\vskip 0.22in
\centering
\includegraphics[width=\figwidth\textwidth]{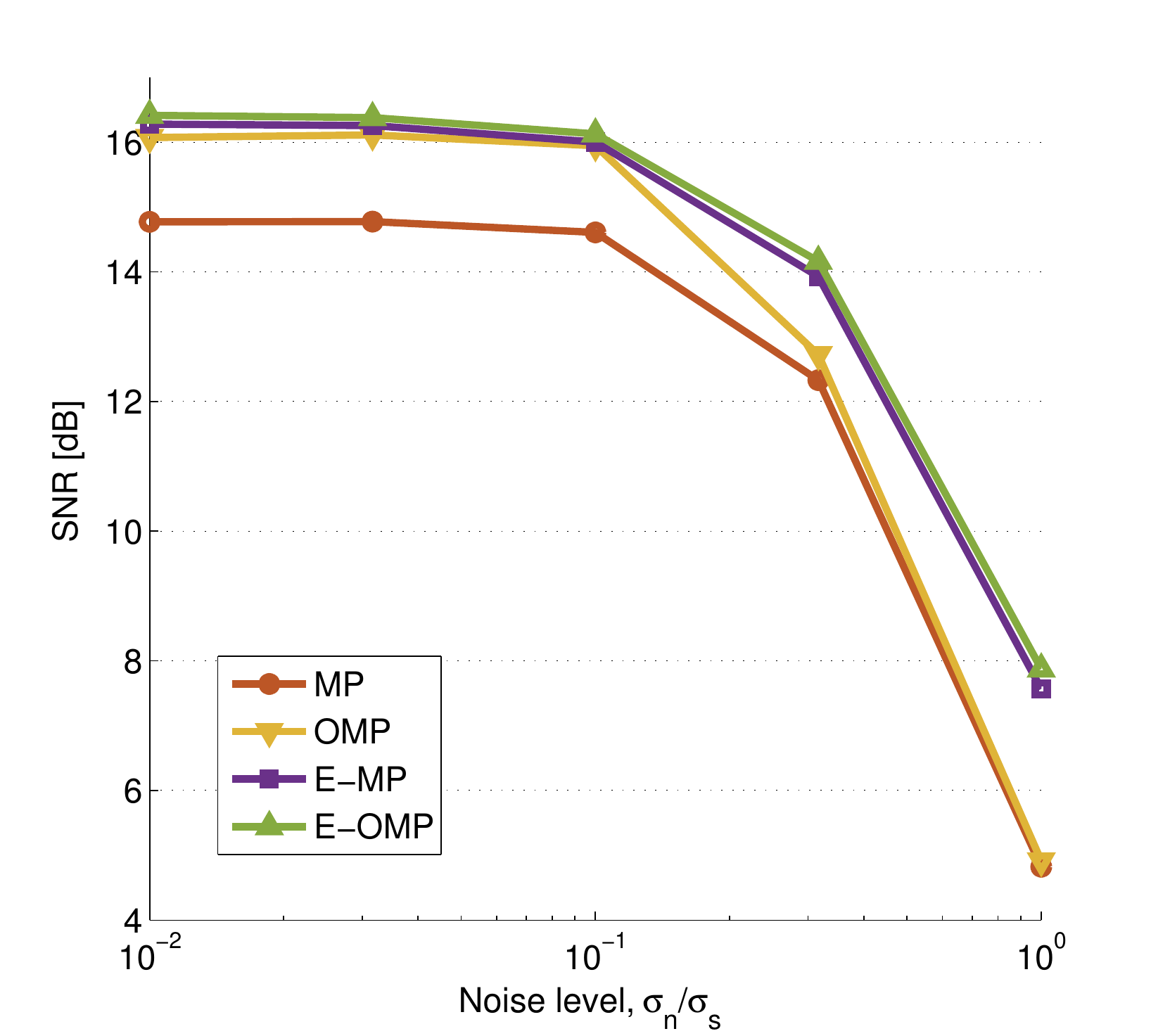}
\caption{{
Reconstruction SNR versus the relative standard deviation of additive Gaussian noise, $\sigma_n / \sigma_s$, for a track of 44.1 kHz rock music with standard deviation $\sigma_s$.}}
% Dictionaries learned from 25 minutes of rock music
% with an average event probability of  0.05.}}
% Result for 44.1kHz audio data (song1.mp3 excluding 10s tails).
% Dictionaries trained on 1500s of 44.1kHz audio (song1.mp3).
\label{fig:denoising}
\end{figure}

Sparse approximations of signals based on learned dictionaries are useful for solving denoising problems because atoms represent repeating additive structure, not noise-like components of the signal which mostly end up in the model residual, $\epsilon(t)$.
We investigate the denoising capability of E--MP- and E--OMP-based models by adding Gaussian noise to the rock music signal and comparing the SNR of the model for different noise levels with $p=0.05$, see \fig{denoising}.
The noise level is quantified in terms of the ratio of the standard deviation of additive Gaussian noise, $\sigma_n$, to the standard deviation of the signal, $\sigma_s$.
As expected the OMP-based model produces denoising results that are far better than MP,
at least for moderate levels of noise, $\sigma_n / \sigma_s \leq 0.1$.
The denoising accuracy obtained with E--MP and E--OMP is comparable to that of OMP for moderate levels of noise, $\sigma_n / \sigma_s \leq 0.1$, and it is superior to OMP for high levels of noise $\sigma_n / \sigma_s > 0.1$.
This is another remarkable consequence of equiprobable atom selection,
which indicates that the selection constraint prevents overfitting of noise-like signal components.

%%%%%%%%%%%%%%%%%%%%%%%%%%%%%%%%%%%%%%%%%%%%%%%%%%%%%%%%%%%%%%%%%%%%%%%%%%%%

\subsection{Computational cost}

\begin{figure}[ht!]
\centering
\includegraphics[width=\figwidth\textwidth]{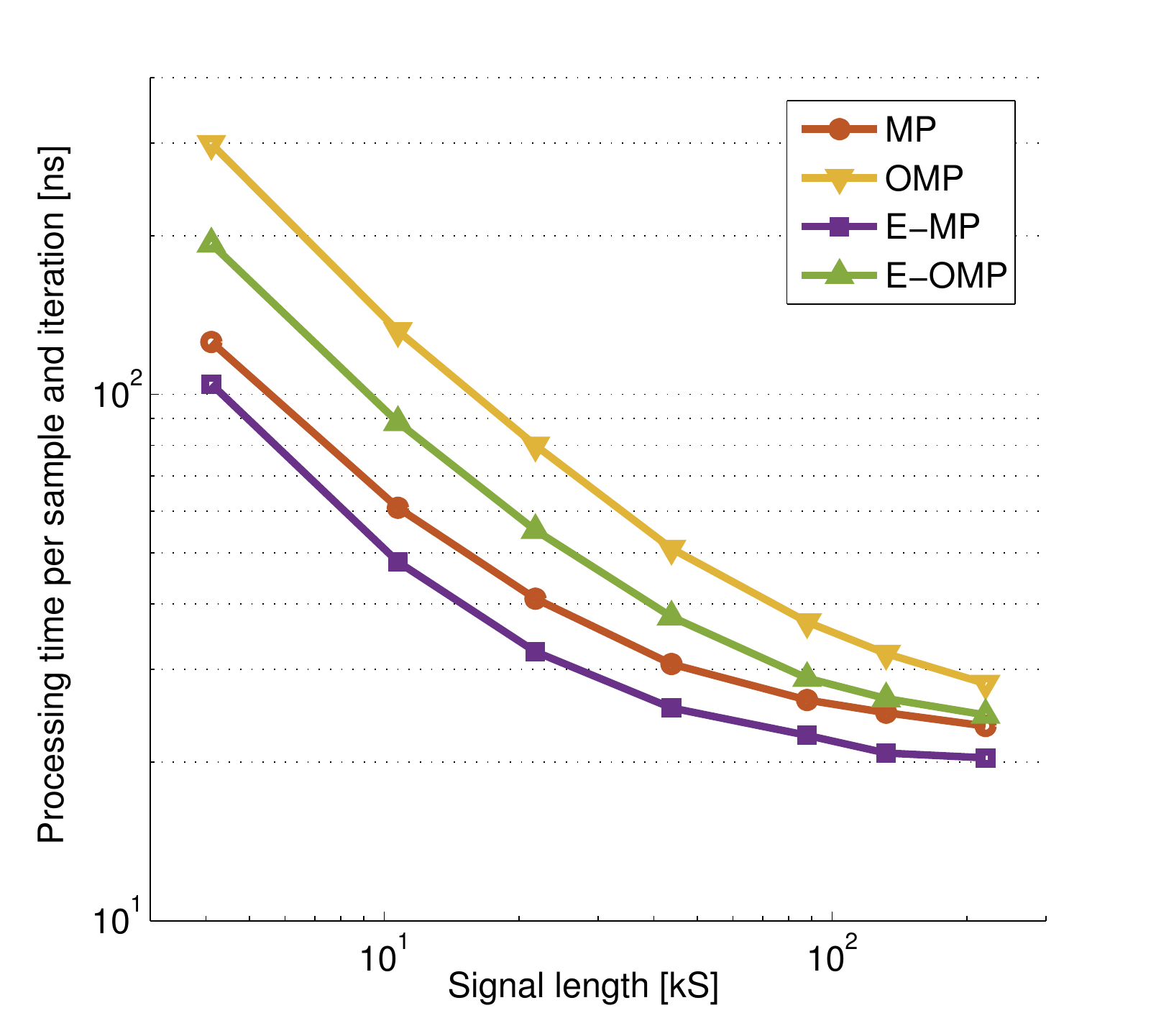}
\caption{{
Processing time versus the window length for dictionaries with 32 atoms using one 2.3 GHz Intel Core i7 processor core.}}
% Dictionaries obtained by training on 3000s of 44.1kHz audio (song1.mp3).
\label{fig:cputime}
\end{figure}

Next we investigate the computational cost of the four different dictionary learning methods.
The experiments are done using one 2.3 GHz Intel Core i7 processor core and a C++ implementation of the algorithms executed in Matlab.
We calculate the average core time per signal sample and iteration of the matching pursuit, see  \fig{cputime}.
This definition implies that the displayed processing time should be multiplied with the number of samples squared times the average atom selection probability, $p$, to get the actual computing time.
We find that equiprobable selection is beneficial in terms of computational cost.
In particular, MP is more costly than E--MP, and OMP is more costly than E--OMP.
Furthermore, with a window length of 200 kilosamples the processing time of E--OMP is comparable to that of MP-based dictionary learning.
Note that these results are obtained with the efficient implementations of MP and local OMP introduced above, which is the reason why the computational cost per sample decreases with the window length.
This is not the case for straightforward implementations of the algorithms, which are significantly more costly.

%%%%%%%%%%%%%%%%%%%%%%%%%%%%%%%%%%%%%%%%%%%%%%%%%%%%%%%%%%%%%%%%%%%%%%%%%%%%

\subsection{Entropy of sparse representation}

Sparse representation with matching pursuit and a learned dictionary is a form of lossy compression where a signal is approximated using prior information encoded in the learned atoms.
Next we study the information entropy of the sparse representations generated by the matching pursuits introduced above and the corresponding learned dictionaries.
The number of selected atoms per second depends on the signal being processed and the method and dictionary used.

\begin{figure}[b!]
\centering
\includegraphics[width=\figwidth\textwidth]{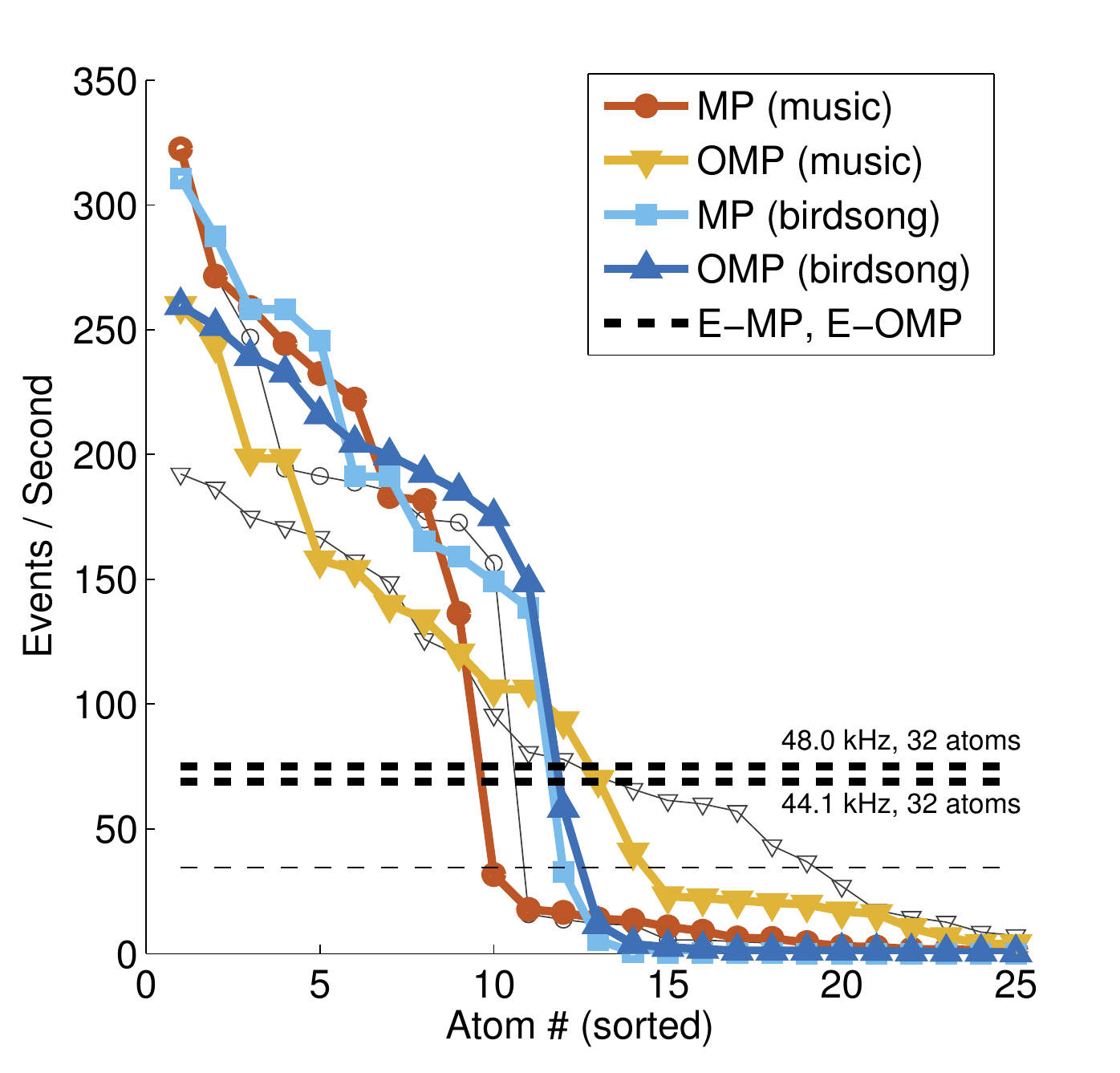}
\caption{Number of atom selection events per second for rock music and birdsong with an average selection probability of $p=0.05$.
Bold (thin) lines are obtained with a dictionary of 32 (64) learned atoms.
For clarity only the twentyfive most frequently selected atoms are included in each case.}
% Average for song1.mp3 after about 3000s of learning.
% 0.05/32 = 0.0016 is the probability in the case of E-MP and E-OMP.
% 0.05*44100/32 = 68.9062
% 0.05*44100/64 = 34.4531
% 0.05*48000/32 = 75
\label{fig:rates}
\end{figure}

The top--25 most frequently MP/OMP-selected atoms obtained using the learned dictionaries are illustrated in \fig{rates} for both rock music and birdsong.
Illustrated in the figure are also the constant rate of atom selection events for E--MP and E--OMP with an average selection probability of $p=0.05$.
For MP and OMP some atoms are typically more likely to be selected than others, while some atoms may not be selected and learned at all.
In contrast to that the average probability for selecting each atom is constant for E--MP and E--OMP.
For example, for the equiprobable pursuits $0.05\times 48000 / 32 = 75$ events per second are expected for each atom on average.
Thus, the entropy of the sequence of selected atom numbers is expected to be higher for E--MP and E--OMP than for the conventional matching pursuits, which is part of the motivation of this study outlined in the Introduction.

We calculate the Shannon entropy, $-\sum_i p_i \text{log}_2(p_i)$, of the atom number sequence generated by the four matching pursuits and the two different signals, see \tab{entropy}.
%
%  MP        OMP       E-MP      E-OMP
%  3.4685    3.5913    5.0000    5.0000	  % Entropy of atom number stream for birdsong with 32 atoms
%  3.5108    4.0669    5.0000    5.0000   % Entropy of atom number stream for rock music with 32 atoms
%  4.6767    5.1631    6.0000    6.0000   % Entropy of atom number stream for rock music with 64 atoms
%
\begin{table}[t!]
\caption{Entropy in bits per atom selected by the different algorithms.}
\begin{center}
\begin{small}
\begin{tabular}{|l|c|cccc|}
\hline
\textsc{Signal} 
& \textsc{No atoms}
& \textsc{MP}
& \textsc{OMP}
& \textsc{E-MP}
& \textsc{E-OMP}
\\
\hline
Birdsong & 32 & 3.5 & 3.6 & 5.0 & 5.0 \\
Music & 32 & 3.5 & 4.1 & 5.0 & 5.0 \\
Music & 64 & 4.7 & 5.2 & 6.0 & 6.0 \\
\hline
\end{tabular}
\end{small}
\end{center}
\vskip -0.1in
\label{tab:entropy}
\end{table}
We find numerically that OMP events have higher entropy than MP events for the signals considered here, and that E--MP and E--OMP have a maximum entropy of $\text{log}_2(M)$ as expected.
For example, with a dictionary of $32$ atoms the entropy is log$_2(32)=5$ bits per selected atom with E--MP and E--OMP.
For comparison, the entropy is $3.5$ ($4.1$) bits per selected atom for MP (OMP) in the case of rock music, and $3.5$ ($3.6$) bits per selected atom for birdsong.

\begin{figure}[t!]
\centering
\includegraphics[width=\figwidth\textwidth]{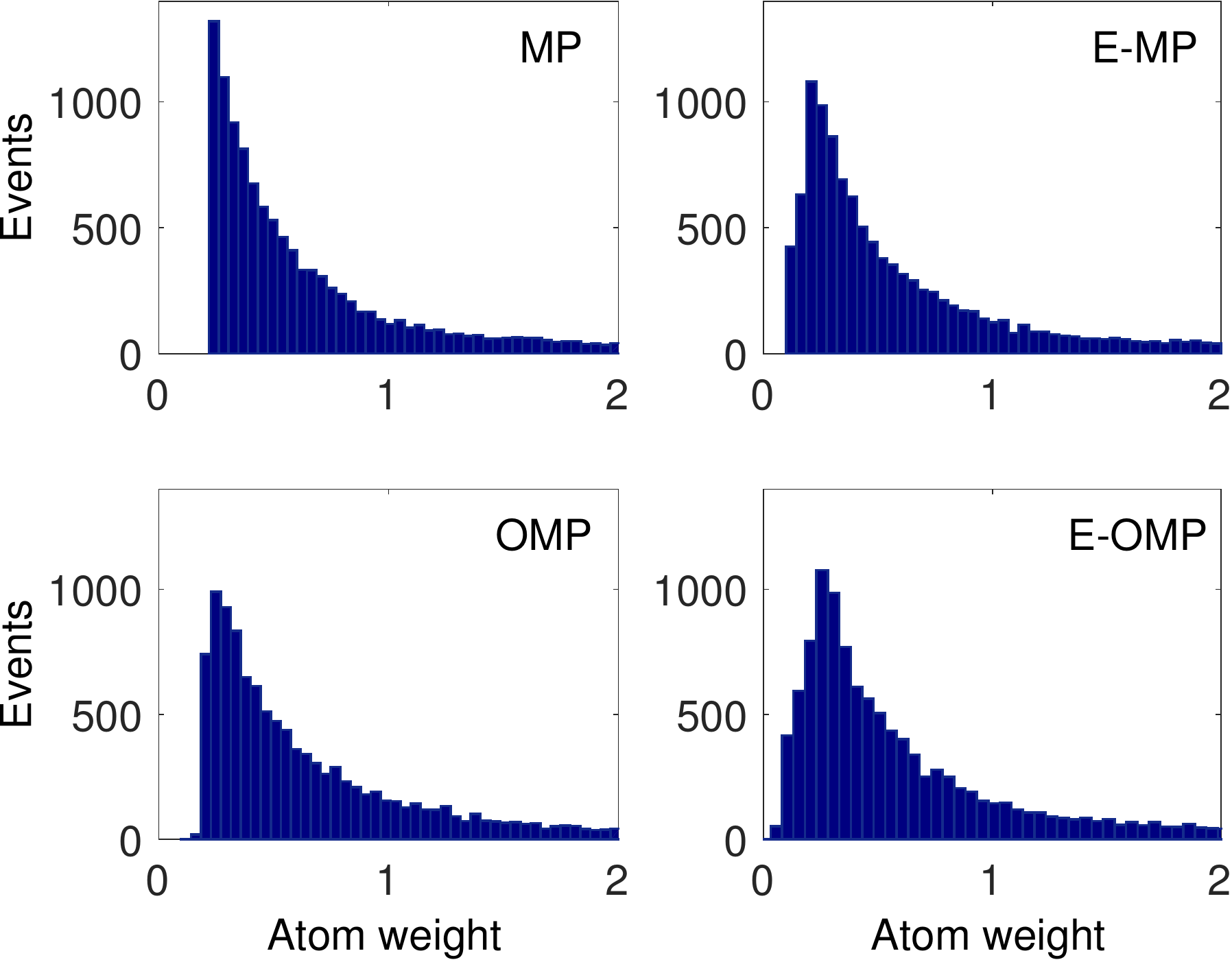}
\caption{Histograms of atom coefficients generated from rock music with 32 learned atoms.}
\label{fig:histogram}
\end{figure}

The distribution of atom coefficients, $a_{m,i}$, also depend on the method used and the signal being processed.
Histograms of the atom coefficients calculated for rock music with 32 learned atoms are illustrated in \fig{histogram}.
An explicit calculation of the Shannon entropy of the event coefficients is not meaningful due to the continuous (floating point) nature of these numbers.
However, we calculate the entropies of multiple histograms with different bin counts (16, 32 and 64) and note a systematic difference between the four models, see \tab{entropy2}.
%
% MP        OMP       E-MP      E-OMP    (numbers normalized with OMP)
% 1.0705    1.0000    1.1019    1.0353   16 bins
% 1.0564    1.0000    1.0760    1.0231   32 bins
% 1.0371    1.0000    1.0495    1.0020   64 bins
%
\begin{table}[t!]
\caption{Coefficient entropies relative to the OMP coefficient entropy.}
\begin{center}
\begin{small}
\begin{tabular}{|c|cccc|}
\hline
\textsc{Bins}
& \textsc{OMP}
& \textsc{E-OMP}
& \textsc{MP}
& \textsc{E-MP}
\\
\hline
16 & 1 & 1.04 & 1.07 & 1.10 \\
32 & 1 & 1.02 & 1.06 & 1.08 \\
64 & 1 & 1.00 & 1.04 & 1.05 \\
\hline
\end{tabular}
\end{small}
\end{center}
\vskip -0.1in
\label{tab:entropy2}
\end{table}
Quantized OMP coefficients have the lowest entropy, followed by E--OMP, MP and E--MP in order of increasing entropy.
The maximum difference is $10$\% when comparing the coefficient entropies of E--MP and OMP with 16 bins (about 4-bits of precision), which can be compared to the difference of about $20$\% in \tab{entropy}.
Thus, by adding the entropies of atom indices and coefficients we conclude that E--MP has the highest event entropy followed by E--OMP for the signals considered here.

In these numerical experiments the average atom selection probability is a constant with value $0.05$,
which implies that on average there is one selected atom in the sparse approximation for every 20 samples of the signal.
This implies that the Shannon information of the sparse approximations generated by the different matching pursuits are directly proportional to the entropy per event.
Therefore, for the signals considered here the sparse approximations calculated with equiprobable selection have higher Shannon information, in line with the higher reconstruction accuracy obtained with these methods in the former subsections.

%%%%%%%%%%%%%%%%%%%%%%%%%%%%%%%%%%%%%%%%%%%%%%%%%%%%%%%%%%%%%%%%%

\section{Discussion}

We extend a well-known dictionary learning and sparse representation model \cite{smith2006} with a basic homeostatic regulation mechanism.
The extension is motivated by the observation that the information entropy of such sparse representations is sub-optimal by construction, and partially also by the central role of homeostatic regulation in cortical networks and spiking neural network models of sensory areas, see for example \cite{King2013,1601.00701}.

The sparse representations are generated with Matching Pursuit (MP) \cite{Mallat1993}, Local Orthogonal MP (OMP) \cite{LocOMP2009} and the homeostatic extensions Equiprobable MP (E--MP) and Equiprobable OMP (E--OMP) introduced here.
Dictionaries of shift-invariant atoms are learned using probabilistic gradient ascent for two different signals (rock music and Zebra Finch song phrases).
With dictionary learning based on E--MP- and E--OMP we obtain an improved rate of convergence and sparse representation SNR (\fig{recsnr}), improved denoising results (\fig{denoising}), a lower computational cost (\fig{cputime}) and higher information entropy of the sparse representation (\tab{entropy} and \tab{entropy2}).

Note that E--MP and E--OMP only make sense in a dictionary learning setting where the atoms are adapted to the signal, otherwise the regular OMP/MP methods should be used.
Furthermore, dictionary learning with equiprobable selection is only sensible with complex signals when the number of shift-invariant atoms in the dictionary is lower than the number of independent and uniqe signal components, which typically is the case in practical applications.
Otherwise the preferred solution is to learn one atom for each independent component, which in general is not expected with equiprobable selection.

The low coefficient entropy of E--OMP in combination with maximum atom selection entropy is an interesting property of E--OMP-based dictionary learning that could improve the accuracy of dictionary learnimg implementations with quantized coefficients.
Furthermore, with E--MP and E--OMP atoms are enforced to occur with the same probability on average and all atoms adapt to the training signal, which is not the case with MP/OMP.
In principle the equiprobable selection mechanism resembles a dropout \cite{srivastava14a} mechanism where the probability of dropout dynamically depends on the selection rate of each atom.
We study the sensitivity of the method to variations in the average selection probability, $p$, and find that the learned dictionaries are useful with other values of $p$ and that the model degrades gracefully for lower values of $p$.
Thus, low-power neuromorphic implementations of the proposed equiprobable selection and dictionary learning methods could be feasible regardless of uncertainties associated with for example device mismatch and noise.

In summary, our main finding is that the accuracy and learning rate of a well-known dictionary learning method are improved by the equiprobable atom selection constraint introduced here, and that the computational cost of the resulting E--MP and E--OMP methods are lower than of MP and local OMP, respectively.
Equiprobable selection can be applied to other greedy methods for dictionary learning and there are also opportunities to further develop the basic homeostatic regulation mechanism outlined here in search for more efficient approaches to address this NP-hard problem.

%%%%%%%%%%%%%%%%%%%%%%%%%%%%%%%%%%%%%%%%%%%%%%%%%%%%%%%%%%%%%%%%%

\section*{Acknowledgment}

This work was stimulated by discussions at the CapoCaccia Neuromorphic Engineering workshops in 2015 and 2016.
In particular F.S. acknowledge the discussions with Prof. Christopher Kello about cortical criticality and the lectures by Prof. Yves Fr\'egnac on receptive fields where a knowledge gap between models and observed properties of cells in the visual cortex were highlighted.
F.S. is funded by a Gunnar \"Oquist Fellowship from the Kempe Foundations and S.M.C. is funded by the SKF--LTU University Technology Center.
We acknowledge travel support from the Swedish Foundation for International Cooperation in Research and Higher Education (STINT), grant number IG2011-2025.

\bibliographystyle{IEEEtran}
\bibliography{main}

\end{document}